\documentclass[runningheads]{llncs}
\usepackage{graphicx}
\usepackage{booktabs}
\usepackage{multirow}
\usepackage{amsmath}
\usepackage{hyperref}
\usepackage{amssymb}
\usepackage{wrapfig}
\usepackage[table]{xcolor}
\usepackage[misc]{ifsym}
\begin{document}
\title{Knowing You at First Glance: Inferring Apparent Personality from Faces}
\author{Shuhuan Chen\inst{1,2,3} \and Xiangyu Zhu\inst{1,4}\textsuperscript{(\Letter)} \and Weisong Zhao\inst{2,3} \and Haichao Shi\inst{2} \and \\ Xiao-Yu Zhang\inst{2}\textsuperscript{(\Letter)} \and Zhen Lei\inst{1,4,5,6}}
\authorrunning{Chen et al.}
%
\institute{
	CBSR\&MAIS, Institute of Automation, CAS, Beijing, China \email{\{xiangyu.zhu,zlei\}@ia.ac.cn} \\
	\and Institute of Information Engineering, CAS, Beijing, China \email{\{chenshuhuan,zhaoweisong,shihaichao,zhangxiaoyu\}@iie.ac.cn} \\
	\and School of Cyber Security, UCAS, Beijing, China \\
	\and School of Artificial Intelligence, UCAS, Beijing, China \\
	\and Centre for Artificial Intelligence and Robotics, Hong Kong Institute of Science \& Innovation, CAS, Hong Kong, China \\
	\and School of Computer Science and Engineering, the Faculty of Innovation Engineering, M.U.S.T, Macau, China
}

\maketitle              
\begin{abstract}
Inferring apparent personality from facial images is important in social scenarios for embodied agents in human-robot interaction. Unlike inferring intrinsic personality traits via conversation, this task models first-impression personality perception based solely on facial appearance before interaction begins. Existing studies mainly focus on the Big Five personality model and often rely on language or multimodal inputs. As a result, it remains unclear whether facial cues alone can support meaningful associations with perceived personality traits. This question is particularly relevant for MBTI types, which are widely used in practice and more readily interpretable by large language models. To this end, we propose \textbf{GlanceFace}, an end-to-end framework for apparent personality inference leveraging vision-language models to introduce semantic priors and a semantic-enhanced facial representation module to capture subtle personality-related cues, together with an uncertainty-aware learning strategy to handle noisy and subjective annotations. Extensive experiments demonstrate strong performance on MBTI-based apparent personality benchmarks and reveal relationships between facial characteristics and perceived personality traits, highlighting its potential to support adaptive initial interaction strategies for embodied agents. The code and dataset are available at \url{https://github.com/MrHuan3/GlanceFace}.

\end{abstract}

\section{Introduction}
Suppose a social interactive robot encounters a user for the first time. Prior to interaction, it has no access to conversational signals and must rely on visual cues to form an initial impression and guide its behavior. This ability to perceive personality before interaction is crucial in real-world social scenarios, where early impressions influence subsequent interaction dynamics. For instance, an agent such as OpenClaw~\cite{openclaw} can construct an initial user profile from this first impression to initialize its dialogue strategy. Despite recent progress in modeling personality-related traits~\cite{Tang_Pan_Zheng_Zhou_Sui_Zhu_Deng_Kuai_2025,SUN2026113255,10.1145/3746270.3760222}, most existing works rely on multimodal inputs~\cite{10536172,10.1007/978-3-319-49409-8_32}, including audio, video, and text, which are typically available only after sufficient interaction. While effective, such approaches are less applicable in early-stage scenarios where only facial images are accessible. In contrast, MBTI personality types~\cite{myers1962myers}, widely used in practice and interpretable by large language models, remain underexplored in this prior-to-interaction setting.

\begin{figure*}[t]
	\centering
	\includegraphics[width=0.95\textwidth]{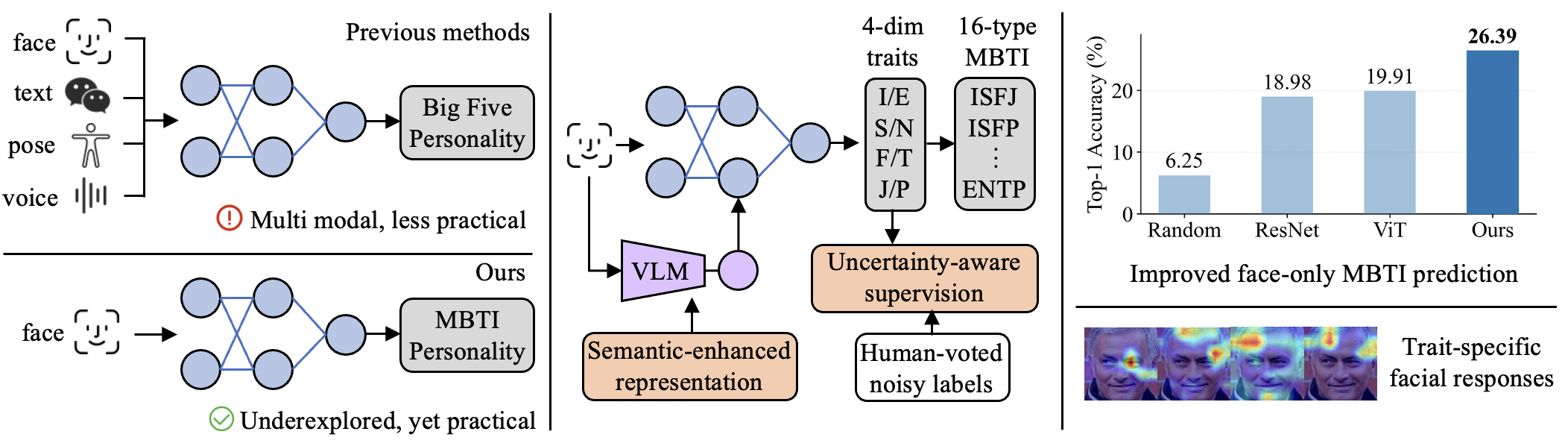}
	\caption{Comparison between previous multimodal Big Five methods and our face-only MBTI framework. We study face-only MBTI prediction with semantic-enhanced representation, uncertainty-aware learning, and interpretable trait-specific facial responses.}
	\label{fig:teaser}
\end{figure*}

This gap raises a fundamental question: \emph{can facial appearance alone provide sufficient cues to learn meaningful associations with perceived MBTI personality traits?} As illustrated in Fig.~\ref{fig:teaser}, we study the practical yet underexplored setting of face-only apparent personality inference. This setting is particularly challenging due to subtle visual cues and subjective annotations. Specifically, the associations between facial characteristics and perceived personality traits are often weak, implicit, and fine-grained, making them difficult to model from static images alone. Meanwhile, personality labels may vary across annotators and contain uncertainty, introducing noisy supervision that degrades prediction accuracy. These challenges call for a model that can effectively exploit subtle semantic facial cues while explicitly accounting for annotation uncertainty.

To this end, we propose \textbf{GlanceFace}, an end-to-end framework for apparent personality inference from facial images. Our framework leverages vision-language models (VLMs) to introduce semantic priors and a semantic-enhanced facial representation module to capture subtle cues associated with perceived MBTI personality traits. To address subjectivity and noise in personality annotations, we further develop an uncertainty-aware learning strategy for reliable training. Extensive experiments demonstrate the effectiveness of GlanceFace on MBTI apparent personality prediction benchmarks, and further analyses provide insights into the associations between facial characteristics and perceived personality traits. Our contributions are summarized as follows:

\begin{itemize}
	\item We present \textbf{GlanceFace}, an end-to-end framework for apparent personality inference from facial images, and study the underexplored prior-to-interaction face-only setting for perceived MBTI personality prediction.
	\item We develop a \textbf{Semantic-Enhanced Facial Representation} module that leverages VLM-derived semantic priors to capture subtle personality-relevant facial cues, along with an \textbf{Uncertainty-Aware Personality Learning} strategy to improve accuracy against subjective and noisy annotations.
	\item Extensive experiments validate the effectiveness of GlanceFace on MBTI-based apparent personality prediction benchmarks, while additional analyses reveal associations between facial cues and perceived personality traits.
\end{itemize}

\section{Related Work}
\subsection{Apparent Personality Inference}
Apparent personality inference aims to model how personality traits are perceived from observable cues and has attracted increasing attention in computer vision~\cite{sun-etal-2024-revealing,Tang_Pan_Zheng_Zhou_Sui_Zhu_Deng_Kuai_2025,10.1007/978-3-319-49409-8_32,8424834}. Most existing studies focus on the Big Five personality model and rely on multimodal inputs, such as facial appearance, voice, and text~\cite{SUN2026113255,Masumura_Orihashi_Ihori_Tanaka_Makishima_Suzuki_Mizuno_Hojo_2025,li2020cr,RYUMINA2024122441}. While these approaches have achieved promising results, their reliance on multimodal signals limits their applicability in scenarios where only facial images are available~\cite{10.1007/978-3-031-22695-3_4}. In contrast to the widely studied Big Five traits, MBTI organizes personality preferences into four binary dimensions: Introversion-Extraversion (I/E), Sensing-Intuition (S/N), Feeling-Thinking (F/T), Judging-Perceiving (J/P), whose combinations define 16 personality types. This formulation provides a structured yet discrete description of personality preferences. Although MBTI-based personality prediction has been investigated in prior work, relatively few studies consider the face-only setting, and direct prediction from facial images remains underexplored~\cite{10724820,Wang_2026_CVPR,zhao2026stavatar}. Therefore, in this work, we focus on the underexplored yet practically important setting of face-only apparent personality inference for perceived MBTI traits.

\subsection{VLMs for Facial Understanding}

VLMs have shown strong capability in learning high-level semantic representations and have been widely applied to visual understanding tasks~\cite{wang2025internvl3_5,liu2023improved,aneja2026phi4reasoningvision15btechnicalreport}. Compared with conventional facial features, VLM-based semantic priors are more effective at capturing subtle and abstract cues~\cite{talon2025seeing,wu2024deepseekvl2mixtureofexpertsvisionlanguagemodels,li2026devilleakagedisentangleddualpurification}. However, their potential for apparent personality inference from facial images remains largely underexplored. This problem is challenging due to weak and implicit associations between facial appearance and perceived personality traits, as well as the inherent subjectivity and noise in personality annotations. Motivated by these challenges, we leverage VLMs to enhance facial representations and incorporate uncertainty-aware learning to better handle noisy supervision during training.

\section{Methods}

\begin{figure*}[t]
	\centering
	\includegraphics[width=0.95\textwidth]{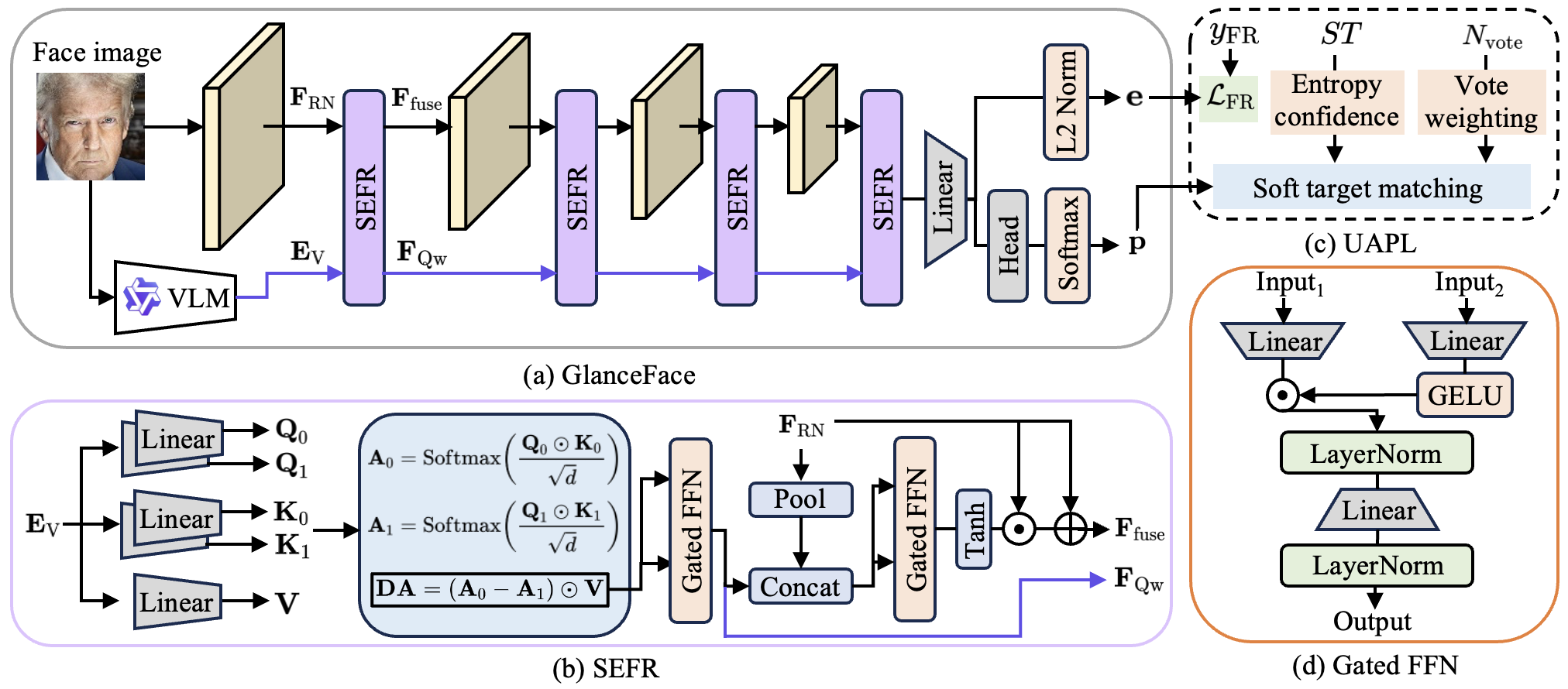}
	\caption{\textbf{Overview of GlanceFace.} (a) The overall framework fuses VLM-derived semantic embeddings with facial features for apparent personality inference. (b) SEFR refines VLM-derived semantic cues through differential gated attention and progressively injects them into hierarchical facial representations. (c) UAPL performs uncertainty-aware supervision by combining soft-target matching, entropy-based confidence, vote-based weighting, and the face recognition loss $\mathcal{L}_{\textrm{FR}}$. (d) The gated FFN enhances informative responses while suppressing noisy activations. Here, $\odot$ denotes element-wise multiplication, and $\oplus$ denotes residual addition.}
	\label{fig:framework}
\end{figure*}

Our framework, termed \textbf{GlanceFace}, is illustrated in Fig.~\ref{fig:framework}. Given a face image, we first employ a modified ResNet backbone~\cite{Deng_2019_CVPR} to extract hierarchical facial features, while a VLM produces semantic embeddings as high-level priors. The semantic embeddings are progressively injected into the ResNet features through the proposed \textit{Semantic-Enhanced Facial Representation} (SEFR) module, leading to enriched facial representations for apparent personality inference. Based on the final fused feature, we use a linear layer and a classification head to predict personality logits, together with an L2-normalized embedding branch for identity-preserving representation learning.

All face images are detected and aligned using MTCNN~\cite{7553523}. To preserve consistency across different images of the same identity, we follow margin-based face recognition methods~\cite{Deng_2019_CVPR,Wang_2018_CVPR,10.1145/3581783.3611711} and encourage identity-consistent facial embeddings, which are optimized with the AdaFace loss~\cite{Kim_2022_CVPR}. Meanwhile, personality prediction is supervised by the proposed \textit{Uncertainty-Aware Personality Learning} (UAPL), which leverages soft labels and vote-based confidence cues for reliable learning under noisy and subjective annotations. The detailed designs of SEFR, UAPL, and the gated FFN are shown in Fig.~\ref{fig:framework}(b)-(d).

\subsection{Semantic-Enhanced Facial Representation}

Recent VLMs~\cite{wang2025internvl3_5,liu2023improved,aneja2026phi4reasoningvision15btechnicalreport} provide rich semantic priors for cross-modal understanding. Motivated by this capability, we enhance facial representation learning by incorporating VLM-derived semantic embeddings for apparent personality inference. Considering both effectiveness and efficiency, we adopt Qwen3-VL-Embedding-2B~\cite{qwen3vlembedding} to extract a semantic embedding from each face image. Compared with conventional face recognition features, these semantic embeddings capture higher-level cues beyond identity-related information, which may be informative for perceived personality understanding. Nevertheless, such semantic priors may also contain noise and uncertainty. To selectively filter and integrate them into facial representations, we propose the SEFR module.

Given a VLM embedding $\mathbf{E}_{\textrm{V}} \in \mathbb{R}^{D_{\textrm{V}}}$, directly applying standard attention is not appropriate, not only because $\mathbf{E}_{\textrm{V}}$ is a global one-dimensional embedding rather than a token sequence, but also because it may contain substantial noise and task-irrelevant semantic components. Inspired by the differential idea in~\cite{ye2025differential}, but tailored to the non-sequential and noisy nature of VLM embeddings in our setting, we design a lightweight differential gated attention mechanism based on element-wise interactions. Specifically, we first project $\mathbf{E}_{\textrm{V}}$ into
\[
\mathbf{Q}_0, \mathbf{Q}_1, \mathbf{K}_0, \mathbf{K}_1, \mathbf{V} \in \mathbb{R}^{d},
\]
and compute the differential semantic response as

\begin{equation}
	\label{eq:diffattn}
	\begin{aligned}
		\mathbf{DA} = \underbrace{\left(\mathrm{Softmax}\left(\frac{\mathbf{Q}_0 \odot \mathbf{K}_0}{\sqrt{d}}\right) - \mathrm{Softmax}\left(\frac{\mathbf{Q}_1 \odot \mathbf{K}_1}{\sqrt{d}}\right)\right)}_{\mathbf{A}_{\textrm{diff}}} \odot \mathbf{V},
	\end{aligned}
\end{equation}

\noindent where $\odot$ denotes element-wise multiplication. Since the projected representations are one-dimensional vectors, we use element-wise query-key interactions for efficient semantic estimation. The differential response $\mathbf{A}_{\textrm{diff}}$ suppresses shared or less informative activations while emphasizing more discriminative semantic cues, and the resulting feature $\mathbf{DA}$ is further modulated by $\mathbf{V}$.

Although the differential response alleviates part of the noise, the resulting semantic interactions may still be entangled. Inspired by gated feed-forward designs~\cite{deepseekai2025deepseekr1incentivizingreasoningcapability}, we further refine $\mathbf{DA}$ through a gated FFN:

\begin{equation}
	\label{eq:gateffn}
	\begin{aligned}
		\mathbf{F}_{\textrm{mid}} &= \mathrm{LN}\!\left( \mathbf{W}_{\textrm{down}} \mathbf{DA} \odot \mathrm{GELU}(\mathbf{W}_{\textrm{gate}} \mathbf{DA}) \right), \\
		\mathbf{F}_{\textrm{Qw}} &= \mathrm{LN}\!\left( \mathbf{W}_{\textrm{up}} \mathbf{F}_{\textrm{mid}} \right),
	\end{aligned}
\end{equation}

\noindent where $\mathrm{LN}$ denotes layer normalization, $\mathrm{GELU}$ denotes the GELU activation function, and $\mathbf{W}_{\textrm{down}}$, $\mathbf{W}_{\textrm{gate}}$, and $\mathbf{W}_{\textrm{up}}$ are learnable linear projection matrices. This gated refinement adaptively enhances informative semantic channels while suppressing residual noisy responses.

We insert one SEFR module after each ResNet block to progressively inject semantic priors into facial representations. Let $\mathbf{F}_{\textrm{RN}}^{i} \in \mathbb{R}^{C \times H_i \times W_i}$ denote the feature map produced by the $i$-th ResNet block. We first apply global average pooling to obtain $\mathbf{F}_{\textrm{P}}^{i} \in \mathbb{R}^{C}$, which is then concatenated with $\mathbf{F}_{\textrm{Qw}}^{i}$ to form $\mathbf{F}_{\textrm{cat}}^{i}$. Another gated FFN transforms $\mathbf{F}_{\textrm{cat}}^{i}$ into a channel-wise modulation vector $\mathbf{F}_{\textrm{g}}^{i}$, which is fused with the ResNet feature as

\begin{equation}
	\label{eq:fuse}
	\begin{aligned}
		\mathbf{F}_{\textrm{fuse}}^{i} = \mathbf{F}_{\textrm{RN}}^{i} + \mathbf{F}_{\textrm{RN}}^{i} \odot \mathrm{Tanh}(\mathbf{F}_{\textrm{g}}^{i}),
	\end{aligned}
\end{equation}

\noindent where $\mathrm{Tanh}$ denotes the hyperbolic tangent activation function, and $\mathbf{F}_{\textrm{g}}^{i}$ is broadcast along the spatial dimensions. The fused feature $\mathbf{F}_{\textrm{fuse}}^{i}$ is then fed into the next ResNet block for further representation learning. Rather than directly concatenating VLM embeddings with facial features, SEFR performs noise-aware semantic refinement and progressively injects the refined semantic priors into hierarchical facial representations, leading to more stable facial representations.

\subsection{Uncertainty-Aware Personality Learning}

A key challenge in apparent personality prediction arises from the noisy and subjective nature of personality annotations. To address this issue, we propose UAPL, which leverages the soft personality distributions and vote statistics provided by the dataset to construct more reliable supervision.

For each sample, the dataset provides a soft distribution for each MBTI dimension, denoted by $ST^{i}(t) \in \mathbb{R}^{2}$, where $t \in \{\mathrm{I/E}, \mathrm{S/N}, \mathrm{F/T}, \mathrm{J/P}\}$ and $i$ indexes the sample. Each sample is also associated with a vote count $N_{\mathrm{vote}}^{i}$, which serves as an indicator of annotation reliability. Intuitively, samples receiving more votes tend to be more reliable. Accordingly, we define the vote-based reliability weight for the $i$-th sample as

\begin{equation}
	\label{eq:vote}
	\begin{aligned}
		w_{\mathrm{vote}}^{i} = \frac{\log(1 + N_{\mathrm{vote}}^{i})} {\frac{1}{B}\sum_{j=1}^{B}\log(1 + N_{\mathrm{vote}}^{j}) + \epsilon},
	\end{aligned}
\end{equation}

\noindent where $B$ denotes the batch size and $\epsilon$ is a small constant for numerical stability. However, vote count alone is insufficient, since two samples with the same number of votes may still differ substantially in label certainty. For instance, a sample with a $95{:}5$ vote distribution is more informative than one with a $55{:}45$ distribution. To quantify this uncertainty, we further define an entropy-based confidence weight for the $t$-th dimension of the $i$-th sample:

\begin{equation}
	\label{eq:entropy}
	\begin{aligned}
		w_{\mathrm{ent}}^{i}(t) = 1 - \frac{1}{\log 2} \left(-\sum_{c=1}^{2} ST_{c}^{i}(t)\log\!\big(ST_{c}^{i}(t)+\epsilon\big) \right).
	\end{aligned}
\end{equation}

This term is the complement of normalized entropy, assigning larger weights to samples with sharper personality distributions. To further reduce the effect of ambiguous supervision, we introduce a confidence indicator:

\begin{equation}
	\label{eq:th}
	\begin{aligned}
		s_{t}^{i} = \begin{cases}
			\max\!\big(ST_0^i(t), ST_1^i(t)\big), & \text{if } \max\!\big(ST_0^i(t), ST_1^i(t)\big) > \tau,\\
			0, & \text{otherwise},
		\end{cases}
	\end{aligned}
\end{equation}

\noindent where $\tau=0.75$ is a predefined threshold based on the dataset. This indicator discards highly ambiguous samples and preserves only those with sufficiently confident personality annotations. Combining the above factors, we define the final uncertainty-aware weight as

\begin{equation}
	\label{eq:finalweight}
	\begin{aligned}
		w^{i}(t) = s_{t}^{i}\, \big(w_{\mathrm{vote}}^{i}\big)^{\alpha} \big(w_{\mathrm{ent}}^{i}(t)\big)^{\beta},
	\end{aligned}
\end{equation}

\noindent where $\alpha$ and $\beta$ control the contributions of the vote-based reliability and entropy-based confidence, respectively. Let $\mathbf{z}^{i}(t)\in\mathbb{R}^{2}$ denote the predicted logits for the $t$-th MBTI dimension of the $i$-th sample, and let

\begin{equation}
	\label{eq:logit2prob}
	\begin{aligned}
		\mathbf{p}^{i}(t)=\mathrm{Softmax}\big(\mathbf{z}^{i}(t)\big)
	\end{aligned}
\end{equation}

\noindent denote the corresponding predicted probability distribution. We then match $\mathbf{p}^{i}(t)$ to the soft target $ST^{i}(t)$ using the Jensen-Shannon divergence:

\begin{equation}
	\label{eq:jsd}
	\begin{aligned}
		\mathrm{JS}\big(\mathbf{p}^{i}(t)\,\|\,ST^{i}(t)\big) = \frac{\mathrm{KL}\big(\mathbf{p}^{i}(t)\,\|\,\mathbf{M}^{i}(t)\big) + \mathrm{KL}\big(ST^{i}(t)\,\|\,\mathbf{M}^{i}(t)\big)}{2},
	\end{aligned}
\end{equation}

where

\begin{equation}
	\begin{aligned}
		\mathbf{M}^{i}(t)=\frac{\mathbf{p}^{i}(t)+ST^{i}(t)}{2}.
	\end{aligned}
\end{equation}

In addition, we encourage the predicted distribution to more closely match the soft target through a margin term:

\begin{equation}
	\label{eq:margin}
	\begin{aligned}
		m^{i}(t)=\left\|\mathbf{p}^{i}(t)-ST^{i}(t)\right\|_{1}^{\gamma},
	\end{aligned}
\end{equation}

\noindent where $\gamma$ controls the penalty strength. The overall UAPL loss is defined as

\begin{equation}
	\label{eq:loss}
	\begin{aligned}
		\mathcal{L}_{\mathrm{UAPL}} = &\frac{\sum_{i=1}^{B}\sum_{t} w^{i}(t)\left[\mathrm{JS}\big(\mathbf{p}^{i}(t)\,\|\,ST^{i}(t)\big) + \lambda\, m^{i}(t)\right]}{\sum_{i=1}^{B}\sum_{t} w^{i}(t)+\epsilon} \\
		&+ \lambda_{\mathrm{FR}}\mathcal{L}_{\mathrm{FR}}(\mathbf{e}, y_{\mathrm{FR}}),
	\end{aligned}
\end{equation}

\noindent where $\lambda$ denotes the margin coefficient, $\lambda_{\mathrm{FR}}$ controls the face recognition regularization term, $\mathcal{L}_{\mathrm{FR}}$ is the AdaFace loss~\cite{Kim_2022_CVPR}, and $\mathbf{e}$ is the facial embedding extracted by the ResNet backbone. In this way, UAPL places greater emphasis on reliable and confident personality annotations while reducing the influence of noisy or ambiguous supervision. As shown in Fig.~\ref{fig:framework}(c), UAPL integrates vote-based weighting, entropy-based confidence, and soft-target matching to provide more reliable supervision under noisy personality annotations.

\subsection{Optimization Objective and Personality Prediction}

For each of the four MBTI dimensions, we apply a Softmax function to the corresponding logit vector $\mathbf{z}^{(t)}$ to obtain a binary probability distribution $\mathbf{p}^{(t)} \in \mathbb{R}^2$ according to Eq.~\ref{eq:logit2prob}. Following a factorized formulation over the four MBTI dimensions, the probability of the $k$-th MBTI type is approximated as the product of the corresponding marginal probabilities:

\begin{equation}
	\label{eq:mbti}
	\mathbf{p}_k^{(\mathrm{mbti})} = \prod_{d=1}^{4} P\!\left(X_d = x_d^{(k)}\right), \quad k \in \{1, \dots, 16\},
\end{equation}

\noindent where $X_d$ denotes the random variable for the $d$-th MBTI dimension, and $x_d^{(k)}$ denotes the corresponding trait of the $k$-th MBTI type. Each marginal probability $P(X_d = x_d^{(k)})$ is obtained from the corresponding entry of $\mathbf{p}^{(d)}$. Based on the resulting $\mathbf{p}_k^{(\mathrm{mbti})}$, we rank all 16 MBTI types in descending order and report the Top-K predictions as the final output, where $K \in \{1, 3, 5\}$.

\begin{table*}[t]
	\centering
	\caption{Performance comparison of different models on 16-type MBTI classification under person-level and image-level evaluation protocols. Metrics include Top-1/3/5 accuracy, F1-score, and AUC (\%). Best: \textbf{bold}; second-best: \underline{underlined}.}
	\label{tab:mbti16_full}
	\setlength{\tabcolsep}{5pt}
	\renewcommand{\arraystretch}{1.15}
	\resizebox{\linewidth}{!}{
		\begin{tabular}{lcccccccccc}
			\toprule
			\multirow{2}{*}{\textbf{Model}}
			& \multicolumn{5}{c}{\textbf{Person-level}}
			& \multicolumn{5}{c}{\textbf{Image-level}} \\
			\cmidrule(lr){2-6} \cmidrule(lr){7-11}
			& \textbf{Acc\text@1}
			& \textbf{Acc\text@3}
			& \textbf{Acc\text@5}
			& \textbf{F1}
			& \textbf{AUC}
			& \textbf{Acc\text@1}
			& \textbf{Acc\text@3}
			& \textbf{Acc\text@5}
			& \textbf{F1}
			& \textbf{AUC} \\
			\midrule
			IR-18 & 18.98 & 51.39 & 68.52 & 12.84 & 75.09 & 15.31 & 37.83 & 53.90 & \underline{11.84} & 63.95 \\
			IR-50 & 20.37 & 50.93 & 70.83 & 12.05 & 70.69 & 15.20 & 38.25 & 55.08 & 10.49 & 60.19 \\
			MobileFace & \underline{22.22} & 49.07 & \underline{71.30} & 11.72 & 68.87 & \underline{16.00} & \underline{38.98} & 56.06 & 10.59 & 59.96 \\
			ViT-S & 19.91 & 50.93 & \underline{71.30} & 11.53 & 72.44 & 15.87 & 38.86 & \underline{56.50} & 10.40 & 61.78 \\
			\midrule
			CosFace & 18.98 & 50.46 & 64.81 & 9.86 & 67.55 & 14.80 & 36.60 & 53.24 & 9.61 &  60.26 \\
			ArcFace & 20.83 & 47.69 & 68.52 & 13.11 & 73.06 & 14.36 & 36.35 & 52.95 & 10.56 & 62.26 \\
			AdaFace & 21.30 & \underline{52.31} & 67.13 & \underline{14.28} & \underline{75.78} & 15.53 & 37.58 & 54.02 & 11.12 & \underline{64.31} \\
			\midrule
			\rowcolor{gray!15} \textbf{Ours} & \textbf{26.39} & \textbf{56.48} & \textbf{75.00} & \textbf{17.99} & \textbf{78.19} & \textbf{18.84} & \textbf{43.90} & \textbf{62.54} & \textbf{12.60} & \textbf{68.91} \\
			\bottomrule
		\end{tabular}
	}
\end{table*}

\section{Experiments}

\subsection{Datasets}

The MBTI personality dataset consists of two parts: a facial image set and a personality annotation set. The facial images are collected from four large-scale face datasets, including MS1MV3~\cite{10.1007/978-3-319-46487-9_6}, CelebA~\cite{Liu_2015_ICCV}, VGGFace2~\cite{8373813}, and IMDB-Face~\cite{Wang_2018_ECCV}. The personality annotations are obtained from the Personality Database~\cite{personality}. The training split contains 3,092 identities with 568,675 images, while the evaluation split contains 216 identities with 39,788 images.

\subsection{Experimental Settings}
We use ResNet-18~\cite{He_2016_CVPR}, following the modification in~\cite{Deng_2019_CVPR}, as the backbone network. All face images are cropped and aligned to $112 \times 112$ using the standard FR preprocessing pipeline~\cite{Kim_2022_CVPR}. We employ Qwen3-VL-Embedding-2B~\cite{qwen3vlembedding} as the VLM to extract 2048-dimensional semantic embeddings. To enforce identity consistency, we adopt AdaFace~\cite{Kim_2022_CVPR} as the FR loss. The backbone is initialized with IR-18 weights pretrained on WebFace4M~\cite{Zhu_2021_CVPR}. The model is trained for 10 epochs using SGD with an initial learning rate of 0.1 and a cosine decay schedule. All experiments are conducted on 8 NVIDIA V100 GPUs.

\subsection{Evaluation Metrics}
We evaluate performance at both the person level and the image level. For person-level evaluation, predictions are aggregated across all images of the same identity, while for image-level evaluation, each image is treated as an independent sample. For the 16-class MBTI classification task, we report Top-K accuracy (Acc@K) with $K \in \{1, 3, 5\}$. We further report F1-score and AUC for both person-level and image-level predictions. For the four MBTI dimensions (i.e., I/E, S/N, F/T, and J/P), each formulated as a binary classification task, we report classification accuracy (Acc), F1-score, and AUC.

\begin{table*}[t]
	\centering
	\caption{Person-level comparison of different methods on the four MBTI dimensions (\%). Metrics include accuracy, F1-score, and AUC. Best: \textbf{bold}; second-best: \underline{underlined}.}
	\label{tab:person_dim}
	\setlength{\tabcolsep}{4pt}
	\renewcommand{\arraystretch}{1.15}
	\resizebox{\linewidth}{!}{
		\begin{tabular}{lccccccccccccccc}
			\toprule
			\multirow{2}{*}{\textbf{Model}} 
			& \multicolumn{3}{c}{\textbf{I-E}} 
			& \multicolumn{3}{c}{\textbf{S-N}} 
			& \multicolumn{3}{c}{\textbf{F-T}}
			& \multicolumn{3}{c}{\textbf{J-P}} 
			& \multicolumn{3}{c}{\textbf{Avg.}} \\
			\cmidrule(lr){2-4} \cmidrule(lr){5-7} \cmidrule(lr){8-10} \cmidrule(lr){11-13} \cmidrule(lr){14-16}
			& \textbf{Acc} & \textbf{F1} & \textbf{AUC} & \textbf{Acc} & \textbf{F1} & \textbf{AUC} & \textbf{Acc} & \textbf{F1} & \textbf{AUC} & \textbf{Acc} & \textbf{F1} & \textbf{AUC} & \textbf{Acc} & \textbf{F1} & \textbf{AUC} \\
			\midrule
			IR-18 & \underline{63.43} & \textbf{61.84} & \underline{70.94} & 74.07 & 83.43 & 69.78 & \underline{69.44} & \textbf{67.00} & 71.38 & 68.98 & 79.26 & 70.02 & 68.98 & \underline{72.88} & 70.53 \\
			IR-50 & 61.11 & 58.42 & 69.11 & 75.46 & 85.24 & 70.23 & 68.06 & 63.87 & 72.61 & 69.44 & 78.43 & 71.74 & 68.52 & 71.49 & 70.92 \\
			MobileFace & \textbf{64.81} & \underline{60.82} & 67.98 & \underline{76.39} & \underline{85.71} & 70.50 & 66.20 & 58.29 & 73.70 & 69.91 & 80.00 & 70.10 & \underline{69.33} & 71.21 & 70.57 \\
			ViT-S & \underline{63.43} & 58.64 & 66.97 & 75.93 & 85.31 & \underline{72.83} & 67.59 & 59.77 & 73.05 & 70.37 & 80.25 & 65.45 & \underline{69.33} & 70.99 & 69.58 \\
			\midrule
			CosFace & 61.11 & 56.25 & 66.68 & 73.61 & 83.57 & 63.85 & 65.28 & 57.63 & 71.88 & \underline{72.22} & \underline{82.46} & 65.94 & 68.06 & 69.98 & 67.09 \\
			ArcFace & 62.04 & 55.42 & 67.62 & 69.91 & 80.94 & 65.27 & 68.06 & 61.88 & \underline{74.39} & 68.52 & 77.92 & 70.42 & 67.13 & 69.04 & 69.43 \\
			AdaFace & 62.50 & 57.14 & 70.21 & 73.61 & 83.85 & 70.26 & 68.06 & 61.88 & 73.54 & 71.76 & 81.11 & \underline{71.93} & 68.98 & 71.00 & \underline{71.48} \\
			\midrule
			\rowcolor{gray!15} \textbf{Ours}    & 62.96 & 57.89 & \textbf{71.29} & \textbf{76.85} & \textbf{85.80} & \textbf{73.63} & \textbf{72.69} & \underline{66.29} & \textbf{77.77} & \textbf{73.15} & \textbf{82.74} & \textbf{74.09} & \textbf{71.41} & \textbf{73.18} & \textbf{74.19} \\
			\bottomrule
		\end{tabular}
	}
\end{table*}

\begin{table*}[t]
	\centering
	\caption{Image-level comparison of different methods on the four MBTI dimensions (\%). Metrics include accuracy, F1-score, and AUC. Best: \textbf{bold}; second-best: \underline{underlined}.}
	\label{tab:image_dim}
	\setlength{\tabcolsep}{4pt}
	\renewcommand{\arraystretch}{1.15}
	\resizebox{\linewidth}{!}{
		\begin{tabular}{lccccccccccccccc}
			\toprule
			\multirow{2}{*}{\textbf{Model}} 
			& \multicolumn{3}{c}{\textbf{I-E}} 
			& \multicolumn{3}{c}{\textbf{S-N}} 
			& \multicolumn{3}{c}{\textbf{F-T}} 
			& \multicolumn{3}{c}{\textbf{J-P}} 
			& \multicolumn{3}{c}{\textbf{Avg.}} \\
			\cmidrule(lr){2-4} \cmidrule(lr){5-7} \cmidrule(lr){8-10} \cmidrule(lr){11-13} \cmidrule(lr){14-16}
			& \textbf{Acc} & \textbf{F1} & \textbf{AUC} & \textbf{Acc} & \textbf{F1} & \textbf{AUC} & \textbf{Acc} & \textbf{F1} & \textbf{AUC} & \textbf{Acc} & \textbf{F1} & \textbf{AUC} & \textbf{Acc} & \textbf{F1} & \textbf{AUC} \\
			\midrule
			IR-18 & 57.03 & \textbf{58.63} & 60.38 & 64.03 & 74.59 & \underline{63.40} & 63.38 & \underline{58.39} & \underline{67.39} & 62.25 & 71.11 & 61.94 & 61.67 & 65.68 & \underline{63.28} \\
			IR-50 & 57.01 & \underline{58.34} & 60.07 & 66.88 & 77.96 & 60.19 & 62.62 & 56.72 & 64.92 & 62.25 & 70.66 & 61.87 & 62.19 & 65.92 & 61.76 \\
			MobileFace & 56.78 & 56.73 & 58.97 & 67.32 & 78.34 & 60.19 & \underline{64.15} & 55.41 & 66.69 & 62.63 & 71.98 & 60.80 & 62.72 & 65.62 & 61.66 \\
			ViT-S & \textbf{57.61} & 57.66 & \underline{60.39} & \underline{67.56} & \underline{78.39} & 62.11 & 63.85 & 56.75 & 66.57 & \underline{63.81} & 73.33 & 60.12 & \underline{63.21} & \underline{66.53} & 62.30 \\
			\midrule
			CosFace & 55.73 & 54.76 & 58.39 & 63.77 & 75.26 & 58.02 & 62.25 & 52.73 & 64.77 & 63.76 & \underline{74.38} & 59.96 & 61.38 & 64.28 & 60.29 \\
			ArcFace & 55.65 & 54.52 & 58.58 & 61.84 & 73.27 & 57.64 & 63.70 & 57.62 & 67.00 & 63.30 & 71.79 & \underline{63.79} & 61.12 & 64.30 & 61.75 \\
			AdaFace & 56.42 & 56.50 & 59.46 & 65.16 & 76.38 & 60.36 & 62.96 & 55.67 & 65.71 & 63.78 & 72.73 & 63.46 & 62.08 & 65.32 & 62.25 \\
			\midrule
			\rowcolor{gray!15} \textbf{Ours}    & \underline{57.44} & 58.15 & \textbf{60.47} & \textbf{72.97} & \textbf{82.68} & \textbf{67.02} & \textbf{69.18} & \textbf{61.29} & \textbf{74.05} & \textbf{66.91} & \textbf{76.89} & \textbf{67.03} & \textbf{66.63} & \textbf{69.75} & \textbf{67.14} \\
			\bottomrule
		\end{tabular}
	}
\end{table*}

\subsection{Performance}
We compare GlanceFace with representative baselines, including general face backbones (e.g., ResNet~\cite{He_2016_CVPR}, MobileFaceNet~\cite{10.1007/978-3-319-97909-0_46}, and ViT~\cite{dosovitskiy2021an}) and FR models (e.g., CosFace~\cite{Wang_2018_CVPR}, ArcFace~\cite{Deng_2019_CVPR}, and AdaFace~\cite{Kim_2022_CVPR}). All FR models use IR-18 as the backbone. As shown in Tab.~\ref{tab:mbti16_full}, \ref{tab:person_dim}, and \ref{tab:image_dim}, GlanceFace outperforms the baselines from both holistic and fine-grained perspectives. For the 16-type MBTI classification task, it achieves the best results under both person-level and image-level protocols, with consistent improvements across Top-K accuracy, F1-score, and AUC. Since these metrics capture complementary aspects of prediction quality, namely ranking quality, class-balanced classification performance, and representation separability, the consistent gains indicate that GlanceFace improves overall personality prediction quality rather than optimizing for a single metric.

The dimension-wise results further reveal the source of the improvement. GlanceFace not only achieves the best average performance over the four MBTI dimensions, but also shows particularly clear advantages on more challenging dimensions such as S/N, F/T, and J/P, while remaining competitive on I/E. This pattern suggests that the proposed method is especially effective at capturing subtle and fine-grained facial cues related to apparent personality inference, rather than only benefiting easier cases. Another noteworthy observation is that the improvement is generally more pronounced under the person-level protocol, where multiple images of the same identity are aggregated. This indicates that GlanceFace learns more stable identity-level personality representations instead of relying on incidental variations in individual images. Moreover, the comparison between IR-18 and deeper backbones (e.g., IR-50) shows that increased backbone capacity does not consistently lead to better performance, suggesting that the gain mainly comes from the proposed personality modeling strategy rather than backbone strength alone.

\subsection{Ablation Study}

\begin{wraptable}{r}{0.6\textwidth}
	\vspace{-30pt}
	\centering
	\caption{Ablation study on person-level 16-type MBTI prediction and average four-dimension performance (\%). Best: \textbf{bold}; second-best: \underline{underlined}.}
	\label{tab:abl}
	\setlength{\tabcolsep}{4pt}
	\renewcommand{\arraystretch}{1.15}
	\resizebox{0.95\linewidth}{!}{
		\begin{tabular}{lcccccc}
			\toprule
			\multirow{2}{*}{\textbf{Model}} 
			& \multicolumn{3}{c}{\textbf{16-MBTI}} 
			& \multicolumn{3}{c}{\textbf{4-dim Avg.}} \\
			\cmidrule(lr){2-4} \cmidrule(lr){5-7}
			& \textbf{Acc\text@1} & \textbf{F1} & \textbf{AUC} & \textbf{Acc} & \textbf{F1} & \textbf{AUC} \\
			\midrule
			IR-18 & 18.98 & 12.84 & 75.09 & 68.98 & 72.88 & 70.53 \\
			w/ SEFR & 19.44 & 14.70 & \underline{76.10} & 69.21 & \underline{73.17} & 71.17 \\
			w/ UAPL & \underline{23.61} & \underline{14.88} & 75.53 & \underline{70.49} & 73.01 & \underline{72.81} \\
			\rowcolor{gray!15} \textbf{Ours} & \textbf{26.39} & \textbf{17.99} & \textbf{78.19} & \textbf{71.41} & \textbf{73.18} & \textbf{74.19} \\
			\bottomrule
		\end{tabular}
	}
	\vspace{-10pt}
\end{wraptable}

We conduct ablation experiments to evaluate the contribution of each component in a principled manner. As shown in Tab.~\ref{tab:abl}, both SEFR and UAPL consistently improve upon the IR-18 baseline. SEFR strengthens representation learning via semantic-guided feature recalibration, which emphasizes informative channels while suppressing noisy responses. As a result, it produces more discriminative representations, reflected by consistent gains in F1-score and AUC across both 16-MBTI and 4-dimension evaluations. UAPL, in contrast, addresses label uncertainty by adaptively weighting samples according to their reliability. From an optimization perspective, this reduces the influence of noisy supervision and stabilizes training, leading to notable improvements in accuracy and AUC.

When combined, SEFR and UAPL deliver further gains across all metrics. This indicates that the two components contribute from complementary perspectives: SEFR improves representation quality, whereas UAPL refines the supervision signal during optimization. Their combination therefore yields more reliable and discriminative personality prediction, confirming that both high-quality representations and reliable supervision are essential for this task.

\subsection{Parameter Sensitivity}

\begin{figure*}[t]
	\centering
	\includegraphics[width=\textwidth]{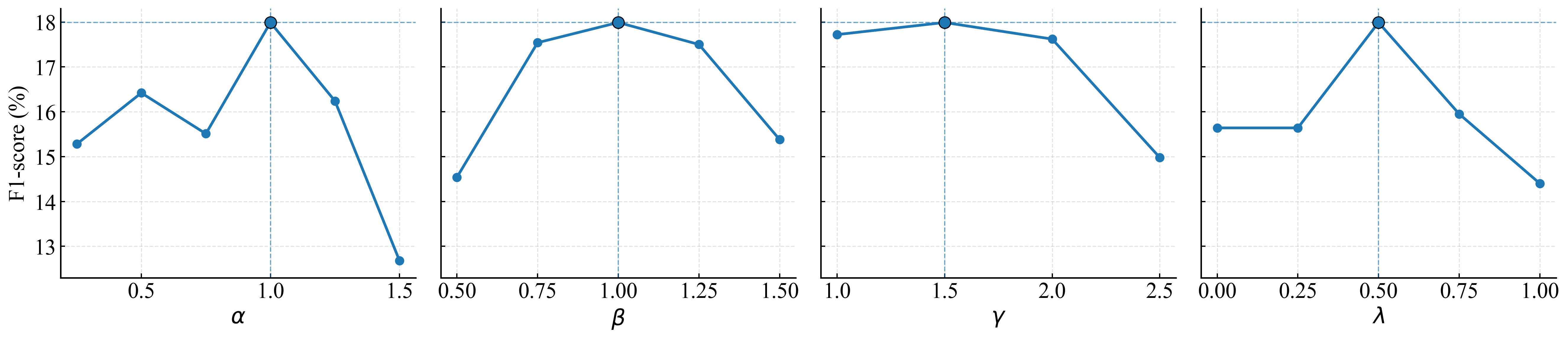}
	\caption{Parameter sensitivity analysis of the UAPL loss with respect to the four hyper-parameters $\alpha$, $\beta$, $\gamma$, and $\lambda$.}
	\label{fig:para_sen}
\end{figure*}

We further analyze the sensitivity of the four hyper-parameters in the UAPL loss, namely the vote-reliability exponent $\alpha$, the entropy-reliability exponent $\beta$, the margin exponent $\gamma$, and the margin weight $\lambda$. As shown in Fig.~\ref{fig:para_sen}, performance is generally stable over a reasonable range of values, while moderate settings consistently produce the best results. Specifically, the optimal performance is achieved at $\alpha=1.0$, $\beta=1.0$, $\gamma=1.5$, and $\lambda=0.5$. This suggests that a balanced weighting strategy is most effective: using vote reliability and target entropy as adaptive sample weights is beneficial, but overly large $\alpha$ or $\beta$ over-emphasizes a small subset of samples and degrades performance. Similarly, a moderate margin setting is preferable. When $\gamma$ or $\lambda$ becomes too large, the margin term dominates the optimization and weakens the distribution-matching objective, resulting in inferior performance. Overall, these results indicate that the proposed loss is reasonably robust to hyper-parameter choices, highlighting the importance of balancing supervision reliability and distribution alignment.

\subsection{Correlation Analysis}

\begin{wrapfigure}{tr}{0.45\textwidth}
	\vspace{-25pt} 
	\centering
	\includegraphics[width=\linewidth]{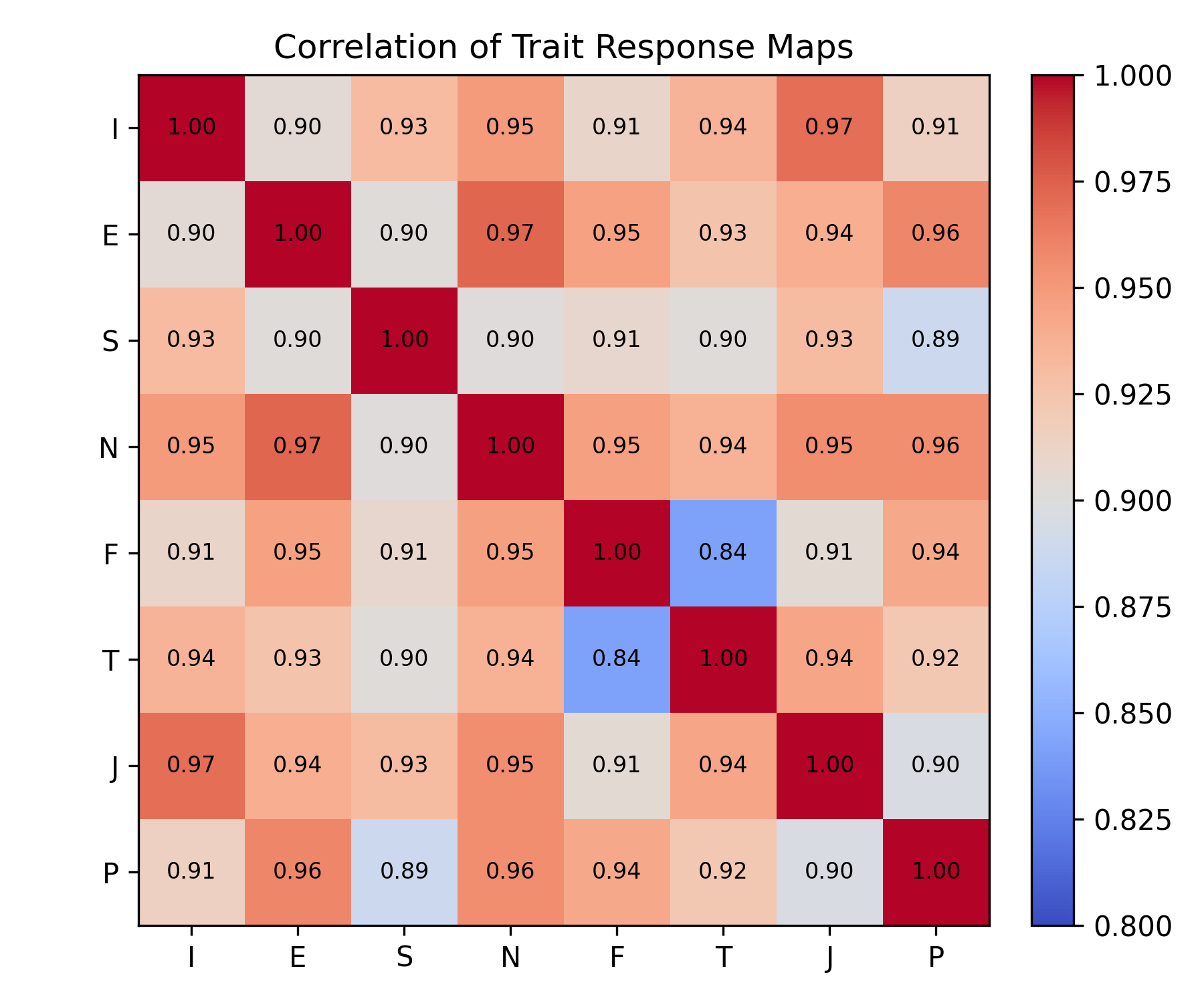}
	\caption{Correlation matrix of trait-specific facial response maps. Opposing traits within the same dimension show lower correlations than cross-dimensional trait pairs.}
	\label{fig:matrix}
	\vspace{-15pt}
\end{wrapfigure}

To better understand the cues learned for MBTI prediction, we analyze trait-specific facial response maps using MTCNN~\cite{7553523} and Grad-CAM++~\cite{8354201}. We aggregate responses to obtain trait-level maps, contrast opposing traits to highlight discriminative regions, and compute pairwise correlations to examine their structure. The results show that most traits share common facial response patterns, while distinctive variations mainly arise between opposing traits within the same dimension, enabling more interpretable comparison of facial cues. This analysis reflects the model’s learned response patterns rather than any deterministic link to real personality traits.

We further examine the correlations among trait-specific facial response maps, as shown in Fig.~\ref{fig:matrix}. Overall, most trait pairs exhibit high correlations, typically above 0.90, suggesting that personality prediction is largely grounded in shared facial regions. Notably, opposing traits within the same dimension tend to show relatively lower correlations than cross-dimensional pairs, with the F-T pair exhibiting the lowest correlation at 0.84. In contrast, many cross-dimensional pairs remain highly correlated, such as I-J and E-N. These observations indicate that personality differences are not driven by distinct spatial regions, but rather by subtle shifts in activation patterns within a shared set of facial cues.

\begin{figure}[t]
	\centering
	\begin{minipage}[t]{0.48\textwidth}
		\centering
		\includegraphics[width=\textwidth]{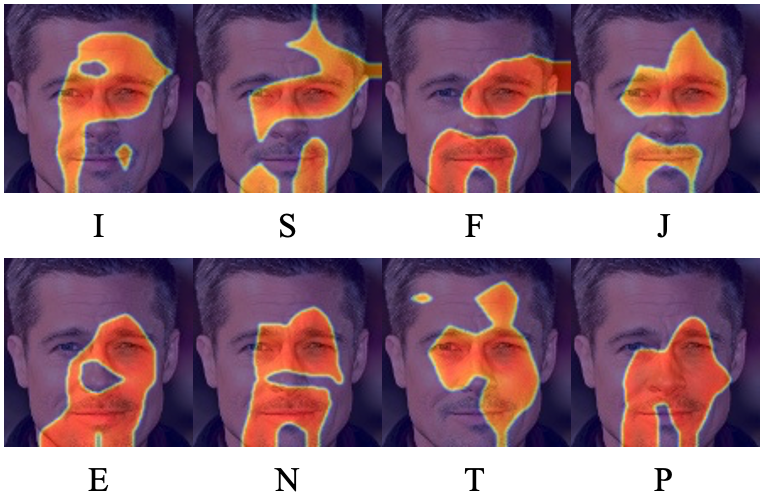}
		\caption{Facial response patterns for different personality traits.}
		\label{fig:final}
	\end{minipage}
	\hfill
	\begin{minipage}[t]{0.48\textwidth}
		\centering
		\includegraphics[width=\textwidth]{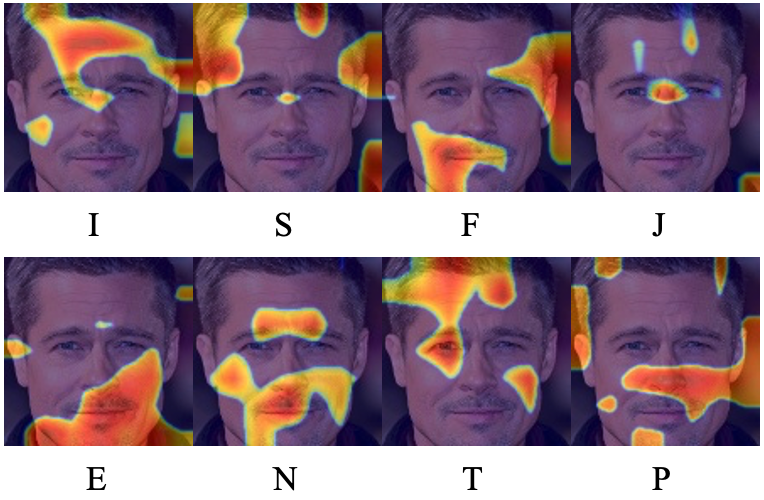}
		\caption{Discriminative regions between opposing personality traits.}
		\label{fig:over}
	\end{minipage}
	\vspace{-10pt}
\end{figure}

As shown in Fig.~\ref{fig:final} and Fig.~\ref{fig:over}, different personality traits exhibit distinct yet largely overlapping facial response patterns, with most activations concentrated on shared regions such as the eyes, nose, and mouth. This indicates that apparent personality inference relies on a common set of facial cues rather than entirely separate spatial regions. The contrastive visualizations further reveal that the differences between opposing traits are mainly reflected as shifts in activation emphasis within these shared areas, rather than distinct region-level separation. In particular, pairs such as F and T show more pronounced variations in activation distribution, suggesting stronger discriminative signals within this dimension. These results demonstrate that personality traits are characterized by structured variations over a shared facial-response basis, where opposing traits are distinguished by subtle but consistent differences in activation patterns.

Overall, these results provide an interpretable view of the model's learned decision basis: it captures stable, trait-dependent facial response patterns with structured correlations across different traits. Importantly, these findings should be regarded as \emph{model-level interpretability evidence} only. They reflect how the trained predictor organizes visual cues for the MBTI classification task, rather than implying any real-world causal or one-to-one relationship between personality traits and facial appearance.

\section{Conclusion}
We make an innovative attempt to study apparent MBTI personality inference from facial images in a first-impression setting prior to interaction, and present \textbf{GlanceFace}. To capture subtle personality-relevant cues, we design a Semantic-Enhanced Facial Representation module that leverages semantic priors from Qwen3-VL and progressively injects them into hierarchical facial representations through a differential gated fusion mechanism. To address the noisy and subjective nature of personality annotations, we further develop an Uncertainty-Aware Personality Learning strategy that emphasizes reliable samples and exploits soft-label supervision for improved learning effectiveness. Extensive experiments demonstrate the effectiveness of GlanceFace at both the person and image levels. Moreover, our analyses offer an interpretable view of the facial response patterns learned for apparent personality prediction. We hope this work will inspire future research on face-based apparent personality understanding.

\section{Broader Impact}
GlanceFace studies apparent personality inference from facial images and may support more adaptive initial interactions for embodied agents and human-computer interfaces. However, the predicted traits reflect perceived impressions rather than intrinsic personality and should not be used for high-stakes decisions. Potential risks include demographic bias, privacy concerns, and misuse in profiling or discrimination. Therefore, the model should be deployed with appropriate consent, fairness evaluation, and human oversight.

\subsubsection{Acknowledgments.}
This work was supported in part by Chinese National Natural Science Foundation Projects 92570119, 62276254, U23B2054, 62376265, 62502514, National Key R\&D Program of China (No. 2025ZD0123501), the Science and Technology Development Fund of Macau Project 0140/2024/AGJ, and InnoHK program.
%
%
%
\bibliographystyle{splncs04}
\bibliography{main}

@String(IJCV  = {Int. J. Comput. Vis.})

@String(CVPR  = {IEEE Conf. Comput. Vis. Pattern Recog.})

@String(ICCV  = {Int. Conf. Comput. Vis.})

@String(ECCV  = {Eur. Conf. Comput. Vis.})

@String(ICLR  = {Int. Conf. Learn. Represent.})

@String(AAAI  = {AAAI})

@String(IJCV  = {IJCV})

@String(CVPR  = {CVPR})

@String(ICCV  = {ICCV})

@String(ECCV  = {ECCV})

@String(ICLR  = {ICLR})

@article{Tang_Pan_Zheng_Zhou_Sui_Zhu_Deng_Kuai_2025, 
	title={Pose as a Modality: A Psychology-Inspired Network for Personality Recognition with a New Multimodal Dataset}, 
	volume={39}, 
	number={2}, 
	journal={AAAI}, 
	author={Tang, Bin and Pan, Ke-Qi and Zheng, Miao and Zhou, Ning and Sui, Jia-Lu and Zhu, Dandan and Deng, Cheng-Long and Kuai, Shu-Guang}, 
	year={2025}, 
	pages={1538-1546}
}

@article{Masumura_Orihashi_Ihori_Tanaka_Makishima_Suzuki_Mizuno_Hojo_2025, 
	title={Multimodal Fine-Grained Apparent Personality Trait Recognition: Joint Modeling of Big Five and Questionnaire Item-level Scores}, 
	volume={39}, 
	number={2}, 
	journal={AAAI}, 
	author={Masumura, Ryo and Orihashi, Shota and Ihori, Mana and Tanaka, Tomohiro and Makishima, Naoki and Suzuki, Satoshi and Mizuno, Saki and Hojo, Nobukatsu}, 
	year={2025}, 
	pages={1456-1464}
}

@book{myers1962myers,
	title={The myers-briggs type indicator},
	author={Myers, Isabel Briggs and others},
	volume={34},
	year={1962},
	publisher={Consulting Psychologists Press Palo Alto, CA}
}

@ARTICLE{8424834,
	author={Gorbova, Jelena and Avots, Egils and Lüsi, Iiris and Fishel, Mark and Escalera, Sergio and Anbarjafari, Gholamreza},
	journal={IEEE MultiMedia}, 
	title={Integrating Vision and Language for First-Impression Personality Analysis}, 
	year={2018},
	volume={25},
	number={2},
	pages={24-33}
}

@ARTICLE{10536172,
	author={Leekha, Maitree and Khan, Shahid Nawaz and Srinivas, Harshita and Shah, Rajiv Ratn and Shukla, Jainendra},
	journal={IEEE Transactions on Affective Computing}, 
	title={VyaktitvaNirdharan: Multimodal Assessment of Personality and Trait Emotional Intelligence}, 
	year={2024},
	volume={15},
	number={4},
	pages={2139-2153}
}

@article{RYUMINA2024122441,
	title = {OCEAN-AI framework with EmoFormer cross-hemiface attention approach for personality traits assessment},
	journal = {Expert Systems with Applications},
	volume = {239},
	pages = {122441},
	year = {2024},
	author = {Elena Ryumina and Maxim Markitantov and Dmitry Ryumin and Alexey Karpov}
}

@article{li2020cr,
	title={Cr-net: A deep classification-regression network for multimodal apparent personality analysis},
	author={Li, Yunan and Wan, Jun and Miao, Qiguang and Escalera, Sergio and Fang, Huijuan and Chen, Huizhou and Qi, Xiangda and Guo, Guodong},
	journal={IJCV},
	volume={128},
	number={12},
	pages={2763--2780},
	year={2020}
}

@InProceedings{10.1007/978-3-319-49409-8_32,
	author={Ponce-L{\'o}pez, V{\'i}ctor
	and Chen, Baiyu
	and Oliu, Marc
	and Corneanu, Ciprian
	and Clap{\'e}s, Albert
	and Guyon, Isabelle
	and Bar{\'o}, Xavier
	and Escalante, Hugo Jair
	and Escalera, Sergio},
	title={ChaLearn LAP 2016: First Round Challenge on First Impressions - Dataset and Results},
	booktitle={ECCV},
	year={2016},
	pages={400--418}
}

@inproceedings{sun-etal-2024-revealing,
	title = {Revealing Personality Traits: A New Benchmark Dataset for Explainable Personality Recognition on Dialogues},
	author = {Sun, Lei  and Zhao, Jinming  and Jin, Qin},
	booktitle = {EMNLP},
	year = {2024},
	pages = {19988--20002}
}

@inproceedings{10.1145/3746270.3760222,
	author = {Niu, Taiyu and Wu, Tianhao and Han, Pengtao and Sun, Shengzhe and Chen, Yifan and Tu, Geng and Xu, Ruifeng},
	title = {A Lightweight Multimodal Framework for Big Five Personality Trait Prediction},
	year = {2025},
	booktitle = {Proceedings of the 3rd International Workshop on Multimodal and Responsible Affective Computing},
	pages = {64–68},
	series = {MRAC '25}
}

@InProceedings{Kim_2022_CVPR,
	author    = {Kim, Minchul and Jain, Anil K. and Liu, Xiaoming},
	title     = {AdaFace: Quality Adaptive Margin for Face Recognition},
	booktitle = {CVPR},
	year      = {2022},
	pages     = {18750-18759}
}

@InProceedings{He_2016_CVPR,
	author = {He, Kaiming and Zhang, Xiangyu and Ren, Shaoqing and Sun, Jian},
	title = {Deep Residual Learning for Image Recognition},
	booktitle = {CVPR},
	year = {2016}
}

@InProceedings{Deng_2019_CVPR,
	author = {Deng, Jiankang and Guo, Jia and Xue, Niannan and Zafeiriou, Stefanos},
	title = {ArcFace: Additive Angular Margin Loss for Deep Face Recognition},
	booktitle = {CVPR},
	year = {2019}
}

@InProceedings{Liu_2015_ICCV,
	author = {Liu, Ziwei and Luo, Ping and Wang, Xiaogang and Tang, Xiaoou},
	title = {Deep Learning Face Attributes in the Wild},
	booktitle = {ICCV},
	year = {2015}
}

@InProceedings{10.1007/978-3-319-46487-9_6,
	author={Guo, Yandong
	and Zhang, Lei
	and Hu, Yuxiao
	and He, Xiaodong
	and Gao, Jianfeng},
	title={MS-Celeb-1M: A Dataset and Benchmark for Large-Scale Face Recognition},
	booktitle={ECCV},
	year={2016},
	pages={87--102},
}

@INPROCEEDINGS{8373813,
	author={Cao, Qiong and Shen, Li and Xie, Weidi and Parkhi, Omkar M. and Zisserman, Andrew},
	booktitle={IEEE FG}, 
	title={VGGFace2: A Dataset for Recognising Faces across Pose and Age}, 
	year={2018},
	volume={},
	number={},
	pages={67-74},
}

@InProceedings{Wang_2018_ECCV,
	author = {Wang, Fei and Chen, Liren and Li, Cheng and Huang, Shiyao and Chen, Yanjie and Qian, Chen and Loy, Chen Change},
	title = {The Devil of Face Recognition is in the Noise},
	booktitle = {ECCV},
	year = {2018}
}

@misc{personality,
	title = {Personality-database},
	howpublished = {\url{https://www.personality-database.com/}},
	note = {Accessed: 2025-12-03}
}

@InProceedings{Wang_2018_CVPR,
	author = {Wang, Hao and Wang, Yitong and Zhou, Zheng and Ji, Xing and Gong, Dihong and Zhou, Jingchao and Li, Zhifeng and Liu, Wei},
	title = {CosFace: Large Margin Cosine Loss for Deep Face Recognition},
	booktitle = {CVPR},
	year = {2018}
}

@inproceedings{10.1145/3581783.3611711,
	author = {Zhao, Weisong and Zhu, Xiangyu and He, Zhixiang and Zhang, Xiao-Yu and Lei, Zhen},
	title = {Cross-Architecture Distillation for Face Recognition},
	year = {2023},
	booktitle = {ACM MM},
	pages = {8076–8085},
}

@article{qwen3vlembedding,
	title={Qwen3-VL-Embedding and Qwen3-VL-Reranker: A Unified Framework for State-of-the-Art Multimodal Retrieval and Ranking},
	author={Li, Mingxin and Zhang, Yanzhao and Long, Dingkun and Chen Keqin and Song, Sibo and Bai, Shuai and Yang, Zhibo and Xie, Pengjun and Yang, An and Liu, Dayiheng and Zhou, Jingren and Lin, Junyang},
	journal={arXiv preprint arXiv:2601.04720},
	year={2026}
}

@article{wang2025internvl3_5,
	title={InternVL3.5: Advancing Open-Source Multimodal Models in Versatility, Reasoning, and Efficiency},
	author={Wang, Weiyun and Gao, Zhangwei and Gu, Lixin and Pu, Hengjun and Cui, Long and Wei, Xingguang and Liu, Zhaoyang and Jing, Linglin and Ye, Shenglong and Shao, Jie and others},
	journal={arXiv preprint arXiv:2508.18265},
	year={2025}
}

@article{liu2023improved,
	title={Improved Baselines with Visual Instruction Tuning}, 
	author={Haotian Liu and Chunyuan Li and Yuheng Li and Yong Jae Lee},
	year={2023},
	journal={arXiv preprint arXiv:2310.03744}
}

@article{deepseekai2025deepseekr1incentivizingreasoningcapability,
	title={DeepSeek-R1: Incentivizing Reasoning Capability in LLMs via Reinforcement Learning}, 
	author={DeepSeek-AI},
	year={2025},
	journal={arXiv preprint arXiv:2501.12948},
}

@ARTICLE{7553523,
	author={Zhang, Kaipeng and Zhang, Zhanpeng and Li, Zhifeng and Qiao, Yu},
	journal={IEEE Signal Processing Letters}, 
	title={Joint Face Detection and Alignment Using Multitask Cascaded Convolutional Networks}, 
	year={2016},
	volume={23},
	number={10},
	pages={1499-1503},
}

@InProceedings{10.1007/978-3-031-22695-3_4,
	author={Gan, Peter Zhuowei
	and Sowmya, Arcot
	and Mohammadi, Gelareh},
	title={Zero-shot Personality Perception From Facial Images},
	booktitle={AI 2022: Advances in Artificial Intelligence},
	year={2022},
	pages={43--56}
}

@InProceedings{10.1007/978-3-319-97909-0_46,
	author={Chen, Sheng
	and Liu, Yang
	and Gao, Xiang
	and Han, Zhen},
	title={MobileFaceNets: Efficient CNNs for Accurate Real-Time Face Verification on Mobile Devices},
	booktitle={Biometric Recognition},
	year={2018},
	pages={428--438}
}

@inproceedings{dosovitskiy2021an,
	title={An Image is Worth 16x16 Words: Transformers for Image Recognition at Scale},
	author={Alexey Dosovitskiy and Lucas Beyer and Alexander Kolesnikov and Dirk Weissenborn and Xiaohua Zhai and Thomas Unterthiner and Mostafa Dehghani and Matthias Minderer and Georg Heigold and Sylvain Gelly and Jakob Uszkoreit and Neil Houlsby},
	booktitle={ICLR},
	year={2021},
}

@article{SUN2026113255,
	title = {From apparent to real: A new path for real personality recognition in robot perception},
	journal = {Pattern Recognition},
	volume = {177},
	pages = {113255},
	year = {2026},
	author = {Yunjia Sun and Shaohui Peng and Tao Wang}
}

@INPROCEEDINGS{10724820,
	author={Prasanna Kumar, R and Bharathi Mohan, G and Rithani, M and Pallavi, Gundala},
	booktitle={ICCCNT}, 
	title={MBTI Based Personality Prediction using Social Media Data}, 
	year={2024},
	volume={},
	number={},
	pages={1-7},
}

@inproceedings{talon2025seeing,
	title={Seeing the Abstract: Translating the Abstract Language for Vision Language Models},
	author={Talon, Davide and Girella, Federico and Liu, Ziyue and Cristani, Marco and Wang, Yiming},
	booktitle={CVPR},
	year={2025}
}

@article{wu2024deepseekvl2mixtureofexpertsvisionlanguagemodels,
	title={DeepSeek-VL2: Mixture-of-Experts Vision-Language Models for Advanced Multimodal Understanding}, 
	author={Zhiyu Wu and Xiaokang Chen and Zizheng Pan and Xingchao Liu and Wen Liu and Damai Dai and Huazuo Gao and Yiyang Ma and Chengyue Wu and Bingxuan Wang and Zhenda Xie and Yu Wu and Kai Hu and Jiawei Wang and Yaofeng Sun and Yukun Li and Yishi Piao and Kang Guan and Aixin Liu and Xin Xie and Yuxiang You and Kai Dong and Xingkai Yu and Haowei Zhang and Liang Zhao and Yisong Wang and Chong Ruan},
	year={2024},
	journal={arXiv preprint arXiv:2412.10302},
}

@InProceedings{Zhu_2021_CVPR,
	author    = {Zhu, Zheng and Huang, Guan and Deng, Jiankang and Ye, Yun and Huang, Junjie and Chen, Xinze and Zhu, Jiagang and Yang, Tian and Lu, Jiwen and Du, Dalong and Zhou, Jie},
	title     = {WebFace260M: A Benchmark Unveiling the Power of Million-Scale Deep Face Recognition},
	booktitle = {CVPR},
	year      = {2021},
	pages     = {10492-10502}
}

@article{aneja2026phi4reasoningvision15btechnicalreport,
	title={Phi-4-reasoning-vision-15B Technical Report}, 
	author={Jyoti Aneja and Michael Harrison and Neel Joshi and Tyler LaBonte and John Langford and Eduardo Salinas},
	year={2026},
	journal={arXiv:2603.03975}
}

@inproceedings{ye2025differential,
	title={Differential Transformer},
	author={Tianzhu Ye and Li Dong and Yuqing Xia and Yutao Sun and Yi Zhu and Gao Huang and Furu Wei},
	booktitle={ICLR},
	year={2025}
}

@INPROCEEDINGS{8354201,
	author={Chattopadhay, Aditya and Sarkar, Anirban and Howlader, Prantik and Balasubramanian, Vineeth N},
	booktitle={WACV}, 
	title={Grad-CAM++: Generalized Gradient-Based Visual Explanations for Deep Convolutional Networks}, 
	year={2018},
	volume={},
	number={},
	pages={839-847},
}

@misc{openclaw,
	title        = {OpenClaw},
	author       = {{OpenClaw Team}},
	howpublished = {\url{https://github.com/openclaw/openclaw}}
}

@InProceedings{Wang_2026_CVPR,
	author    = {Wang, Baiqin and Zhu, Xiangyu and Shen, Fan and Xu, Hao and Lei, Zhen},
	title     = {PC-Talk: Precise Facial Animation Control for Audio-Driven Talking Face Generation},
	booktitle = {Proceedings of the IEEE/CVF Conference on Computer Vision and Pattern Recognition (CVPR)},
	year      = {2026},
	pages     = {25153-25162}
}

@misc{li2026devilleakagedisentangleddualpurification,
	title={The Devil Is in the Leakage: A Disentangled Dual-Purification Framework for High-Fidelity Hairstyle Transfer}, 
	author={Jijie Li and Jiankuo Zhao and Xiangyu Zhu and Zhen Lei},
	year={2026},
	eprint={2607.11281},
	archivePrefix={arXiv},
	primaryClass={cs.CV},
}

@inproceedings{zhao2026stavatar,
	title={Stavatar: Soft binding and temporal density control for monocular 3d head avatars reconstruction},
	author={Zhao, Jiankuo and Zhu, Xiangyu and Wang, Zidu and Lei, Zhen},
	booktitle={Proceedings of the IEEE/CVF Conference on Computer Vision and Pattern Recognition},
	pages={10996--11005},
	year={2026}
}

\end{document}